%% file: main.tex
\documentclass[conference]{IEEEtran}
\IEEEoverridecommandlockouts
\usepackage{amsmath,amssymb,amsfonts,amsthm}
\usepackage{algorithm}
\usepackage{algpseudocode}
\usepackage{graphicx}
\usepackage{booktabs}
\usepackage{multirow}
\usepackage{xcolor}
\usepackage{cite}
\usepackage{url}
\usepackage[hidelinks]{hyperref}

\newtheorem{proposition}{Proposition}

\newcommand{\dC}{\Delta C}
\newcommand{\E}{\mathbb{E}}
\newcommand{\KL}{\mathrm{KL}}
\newcommand{\sg}{\mathrm{sg}}
\newcommand{\method}{ZipDrift}

\begin{document}

\title{\method: Generating Multimodal Short-Horizon\\ MPC Costs with One-Step Drifting Models}

\author{\IEEEauthorblockN{Ke Zhang}
\IEEEauthorblockA{Trinity College Dublin, Dublin, Ireland\\
zhangk2@tcd.ie}}

\maketitle

\begin{abstract}
Short-horizon model predictive control (MPC) is fast but myopic. ZipMPC compensates by learning a context-dependent correction of the short-horizon cost that imitates a long-horizon expert. We observe that the optimal correction is often multimodal: when a slower opponent can be passed on either side, good corrections of opposite sign coexist, and the expert's choice between them hinges on tiny cost differences. Deterministic regression then averages modes or inherits inconsistent expert tie-breaking. We propose \method, which trains a one-step drifting generator over MPC cost corrections, tilted toward low closed-loop cost and learned only from cost evaluations of candidate corrections, without differentiating through the optimiser. A few generated candidates are ranked by a learned critic, and the model-based MPC then enforces dynamics and constraints. In a kinematic overtaking benchmark derived from ZipMPC, more than half of the contexts are bimodal and the manual short-horizon MPC crashes in 28\% of them. With an eight-dimensional stagewise correction, evaluated on 1000 fresh contexts and five training seeds, \method{} with eight candidates crashes in 0.74\% of episodes, compared with 3.4\% for an idealised ZipMPC-style regressor and 2.1\% for critic-ranked random candidates with the same budget. A mixture density network (MDN) with the same selection is equally good (0.76\%). The benefit of drifting lies in sample quality: in a pre-specified test on further unseen contexts, a single drifting sample crashes in 8.7\% of episodes against 13.2\% for the MDN, and \method{} needs four candidates instead of eight to fall below 1\%. With many candidates, all methods degrade as selection exploits errors of the critic.
\end{abstract}


\input{sec_intro}
\input{sec_related}
\input{sec_method}
\input{sec_exp_setup}
\input{sec_exp_results}

\input{sec_discussion}
\input{sec_conclusion}

\bibliographystyle{IEEEtran}
\bibliography{refs}

\end{document}

%% file: sec_intro.tex
\section{Introduction}
\label{sec:intro}

Model predictive control (MPC) is the method of choice for agile vehicles because it enforces dynamics and constraints explicitly, but its computational cost grows with the prediction horizon. Short horizons run in real time yet act myopically: a race car that cannot see past the next few metres brakes late for corners and cannot prepare an overtake. ZipMPC~\cite{rickenbach2025zipmpc} addresses this by compressing a long horizon into the cost of a short one. A neural network maps context, such as the upcoming track curvature, to a correction of the short-horizon cost, and is trained so that the short MPC imitates a long-horizon expert. The learned correction acts like a pseudo reference: on the racing benchmark it shifts the vehicle laterally in anticipation of curves that lie beyond the horizon.

This paper starts from an observation about what such a correction must represent. Consider a slower opponent ahead, outside the short horizon but inside the long one. Passing on the left and passing on the right are both good, and they require lateral cost corrections of opposite sign. The optimal correction is therefore a \emph{set} with two separated components, and which one an expert picks can depend on arbitrarily small differences in cost. A deterministic map trained by regression, as in ZipMPC and related weight-adaptation schemes~\cite{trisovic2026vigor,frohlich2022contextual,zarrouki2026sgrl}, converges to the conditional mean of its labels, which lies between the modes, i.e., straight into the opponent (Proposition~\ref{prop:avg}).

Generative models represent such multimodality naturally, and recent drifting models~\cite{deng2026drifting} do so with a single network evaluation. Drifting MPC~\cite{foffano2026drifting} tilts a drifting generator toward low cost to plan trajectories from offline data, but it does not use a dynamics model, does not enforce constraints, scores plans on generated states, and has so far been evaluated on a linear mass--spring--damper system. Trajectory-space drifting also suffers from a kernel geometry dominated by far-future dimensions, which collapses plans to their average~\cite{puthumanaillam2026kdp}.

We propose \method, which combines the two ideas in the space where each is strongest. A one-step drifting generator, tilted by the closed-loop cost, outputs the cost correction of a model-based short-horizon MPC; the MPC enforces dynamics and constraints; a small number of candidates is generated in one pass and the best is selected. Our contributions are:
\begin{itemize}
\item We identify and quantify mode averaging in context-dependent MPC cost learning. In an overtaking task built on the ZipMPC kinematic racing setup, 53\% of contexts admit two separated sets of good corrections, and the long-horizon expert's two local solutions differ by less than 0.5\% in cost in about half of the contexts where both exist. Imitation is therefore sensitive to how the expert breaks near-ties.
\item We formulate cost-parameter-space drifting with a Gibbs target over MPC cost corrections. Training only needs closed-loop cost \emph{evaluations} of candidate corrections, not gradients through the optimiser, so labels come from the deployment solver with hard constraints.
\item On 1000 fresh test contexts with an eight-dimensional stagewise correction, generated candidates ranked by a critic crash in 0.74\% (\method) and 0.76\% (MDN) of closed-loop episodes, whereas an idealised ZipMPC-style regressor crashes in 3.4\% and critic-ranked random candidates with the same budget in 2.1\%. Drifting does not lower the attainable crash rate relative to the MDN, but its samples are better: in a test fixed before it was run, on contexts not used before, a single drifting sample crashes in 8.7\% of episodes against 13.2\%, and \method{} needs four candidates instead of eight to fall below 1\%. We also report where the generator does not help: in one dimension simpler multimodal baselines are as effective, and with many candidates selection exploits errors of the critic.
\end{itemize}
This is a preliminary study on a deliberately minimal benchmark; Section~\ref{sec:discussion} lists its limitations.

%% file: sec_related.tex
\section{Related Work}
\label{sec:related}

\textbf{Learning costs to shorten MPC horizons.}
Short prediction horizons make nonlinear MPC real-time capable but myopic, and the resulting suboptimality is well understood for unconstrained schemes~\cite{gruene2009horizon}. One remedy is to learn the ingredients of the short-horizon problem. Jain et al.~\cite{jain2021costdesign} learn a surrogate stage cost so that a short-horizon MPC optimises a long-horizon objective. ZipMPC~\cite{rickenbach2025zipmpc} imitates a long-horizon MPC by learning a context-dependent correction of the cost of a short-horizon contouring MPC~\cite{liniger2015mpcc}, propagating gradients of an imitation loss through the optimiser with differentiable MPC~\cite{amos2018diffmpc}; VIGOR-MPC~\cite{trisovic2026vigor} extends the idea to visual context. Weights can also be tuned per context with Bayesian optimisation~\cite{frohlich2022contextual} or adapted by reinforcement learning~\cite{zarrouki2026sgrl,gros2020dempc}, and an actor or critic can supply warm starts and terminal costs~\cite{reiter2024ac4mpc}. Approximate MPC instead replaces the optimiser by a network~\cite{hertneck2018approxmpc}. All of these learn a \emph{deterministic} map from context to cost parameters (or a unimodal Gaussian policy over them). We show that the optimal correction can be genuinely multimodal and that deterministic maps then average modes into infeasible or colliding behaviour.

\textbf{Generative models for planning and MPC.}
Diffusion models generate trajectories~\cite{janner2022diffuser} and action chunks~\cite{chi2023diffusionpolicy} and capture multimodal behaviour. They have been combined with MPC in several ways: as learned dynamics and proposal models~\cite{zhou2024dmpc}, by computing the score from a known model~\cite{pan2024mbd}, as approximate MPC policies that select among sampled modes using the MPC cost~\cite{julbe2025diffampc}, and as reference generators tracked by an MPC on miniature race cars~\cite{ma2025codig}. Large-scale studies indicate that sampling followed by selection is often preferable to gradient guidance~\cite{lu2025diffplanner}. Energy-weighted generative objectives bias samples toward low cost~\cite{zhang2025efm}, echoing the exponential tilting of control as inference~\cite{levine2018controlinference} and MPPI~\cite{williams2017mppi}. These methods generate trajectories or actions; none generates the \emph{cost parameters} of a short-horizon MPC.

\textbf{Drifting models.}
Drifting models~\cite{deng2026drifting} train a one-step generator by regressing its output toward itself plus a kernel mean-shift field that attracts samples to data and repels them from each other, so inference costs a single network evaluation. Drifting MPC~\cite{foffano2026drifting} tilts the attraction field by $e^{-\beta J}$ to sample trajectories from a cost-tilted distribution learned from offline data, without a dynamics model, and selects among a few samples. Keyed drift fields~\cite{puthumanaillam2026kdp} show that trajectory-space drifting collapses to average plans unless neighbourhoods are computed in a compact key space, and drifting policies have been tilted toward high value at the action level~\cite{xia2026dfp}. We move drifting into the low-dimensional cost-parameter space of a model-based short-horizon MPC, where the kernel geometry is benign and the MPC restores dynamic feasibility and constraint satisfaction.

%% file: sec_method.tex
\section{Problem Setting}
\label{sec:problem}

\textbf{Short-horizon MPC with a learned cost correction.}
We consider a discrete-time system $x_{k+1}=f(x_k,u_k)$ controlled by a short-horizon MPC with horizon $N_S$,
\begin{equation}
\label{eq:mpc}
\begin{aligned}
\min_{\{x_i,u_i\}}\ & \sum_{i=0}^{N_S-1} \tfrac12\big\|[x_i;u_i]\big\|^2_{\mathrm{diag}(q)} + (p+\dC)^\top [x_i;u_i]\\
\text{s.t.}\ & x_{i+1}=f(x_i,u_i),\quad x_0=x(k),\quad (x_i,u_i)\in\mathcal{Z}_i,
\end{aligned}
\end{equation}
where $(q,p)$ is a manually tuned cost, $\dC\in\mathcal{Y}\subset\mathbb{R}^{d_c}$ is a correction of the linear cost terms, and $\mathcal{Z}_i$ collects input, track and obstacle constraints. This is exactly the parameterisation of the released ZipMPC implementation, which learns linear cost terms on top of a hand-tuned contouring cost~\cite{rickenbach2025zipmpc}. Let $\pi_{\dC}$ denote the closed-loop controller that applies the first input of~\eqref{eq:mpc}. A context $c$ summarises the information a long horizon would see (in ZipMPC, upcoming track curvature; here, also the state of an opponent).

\textbf{Task cost.}
For a context $c$ and a correction $\dC$, let $J(\dC;c)$ be the cost of running $\pi_{\dC}$ from $c$ for $T_r$ steps (negative progress plus a penalty for crashes). ZipMPC trains a deterministic map $h_\theta(c)$ so that the short MPC imitates a long-horizon MPC; the best such map would output a minimiser of $J(\cdot;c)$ or of an imitation loss with respect to the expert.

\textbf{Mode averaging.}
When the long horizon reveals an obstacle that can be passed on either side, the lateral cost term $p_d$ that pre-positions the vehicle has two good values of opposite sign. Deterministic regression cannot represent this.

\begin{proposition}[Mode averaging]
\label{prop:avg}
Let labels $\dC$ be drawn from a conditional distribution $p(\cdot\mid c)$, e.g., expert-optimal corrections with ties broken arbitrarily. The minimiser of $\E\|h(c)-\dC\|^2$ over all measurable $h$ is $h^\star(c)=\E[\dC\mid c]$. If $p(\cdot\mid c)$ puts mass $\alpha$ and $1-\alpha$ on two separated near-optimal sets $S_1,S_2$ and $J(\cdot;c)$ is large on the segment between them except near its endpoints, then $J(h^\star(c);c)$ is large for all $\alpha$ bounded away from $0$ and $1$.
\end{proposition}

The first claim is the standard characterisation of the conditional mean; the second follows by noting that $h^\star(c)$ lies on the segment between $\E[\dC\mid \dC\in S_1]$ and $\E[\dC\mid \dC\in S_2]$. Section~\ref{sec:exp} shows that this situation is common in a simple overtaking task and that it causes collisions.

\section{\method: Cost-Tilted Drifting in Cost Space}
\label{sec:method}

\textbf{Target distribution.}
Instead of a point estimate we learn a conditional generator $G_\phi(\epsilon,c)\in\mathcal{Y}$, $\epsilon\sim\mathcal{N}(0,I)$, whose output distribution approximates the cost-tilted target
\begin{equation}
\label{eq:target}
p_\beta(\dC\mid c)\ \propto\ p_0(\dC\mid c)\,\exp\!\big(-\beta\,J(\dC;c)\big),
\end{equation}
with a prior $p_0$ over admissible corrections (uniform on a box in our experiments). By the Gibbs variational principle, $p_\beta$ is the unique minimiser of $\E_{q}[J]+\beta^{-1}\KL(q\,\|\,p_0)$ over distributions $q$; Foffano et al.~\cite{foffano2026drifting} use this characterisation for trajectories, and the argument does not depend on how $J$ is evaluated. As $\beta$ grows, $p_\beta$ concentrates on all near-optimal corrections, keeping every mode.

\textbf{Drifting objective.}
Following drifting models~\cite{deng2026drifting} and their cost-tilted variant~\cite{foffano2026drifting}, for a context $c$ we draw $M$ generated samples $y_m=G_\phi(\epsilon_m,c)$ and use $K$ labelled candidates $y_j^+$ with costs $J_j$ as positives. The drift field at $y_m$ is
\begin{align}
V(y_m)&=\sum_{\tau\in\mathcal{T}}\big[A_\tau(y_m)-R_\tau(y_m)\big], \label{eq:drift}\\
A_\tau(y_m)&=\frac{\sum_j w_j k_\tau(y_m,y_j^+)\,(y_j^+-y_m)}{\sum_j w_j k_\tau(y_m,y_j^+)}, \nonumber\\
R_\tau(y_m)&=\frac{\sum_{m'\neq m} k_\tau(y_m,y_{m'})\,(y_{m'}-y_m)}{\sum_{m'\neq m} k_\tau(y_m,y_{m'})}, \nonumber
\end{align}
with Laplace kernels $k_\tau(y,y')=\exp(-\|y-y'\|/\tau)$ at several temperatures $\mathcal{T}$ and tilting weights $w_j\propto\exp(-\beta\,z_j)$, where $z_j$ is the cost $J_j$ standardised within the candidate pool of $c$ and clipped to $[-5,5]$ as in the Drifting MPC implementation. Because the weights enter a self-normalised average, the normalising constant of~\eqref{eq:target} cancels~\cite{foffano2026drifting}. The generator is trained with the drifting loss
\begin{equation}
\label{eq:loss}
\mathcal{L}(\phi)=\E_{c,\epsilon}\big\|G_\phi(\epsilon,c)-\sg\big(G_\phi(\epsilon,c)+V\big)\big\|^2,
\end{equation}
where $\sg$ stops gradients. The field vanishes when the generated distribution matches the tilted positives.

\textbf{Zeroth-order, solver-agnostic labels.}
The only supervision is the scalar cost $J_j$ of candidate corrections, obtained by running the short-horizon MPC in closed loop. No gradient through the optimiser is needed, so the labels can be produced by the same constrained solver used at deployment, by a high-fidelity simulator, or by logged rollouts. This removes three difficulties of differentiable-MPC training reported for ZipMPC: gradients of a local convex approximation, soft state constraints during training, and a mismatch between the training and deployment solvers~\cite{rickenbach2025zipmpc}. Labelling is embarrassingly parallel.

\textbf{Inference with selection.}
At decision time the generator produces $M$ candidates in one batched forward pass. A critic $\hat J_\psi(c,\dC)$, regressed on the same labels, selects $\dC^\star=\arg\min_m \hat J_\psi(c,y_m)$, which is then used in~\eqref{eq:mpc}. Selection among i.i.d.\ samples from an approximation of $p_\beta$ inherits the best-of-$M$ argument of~\cite{foffano2026drifting}: the probability of choosing a $\delta$-suboptimal candidate decays exponentially in $M$ when the generator is close to $p_\beta$ in total variation and $p_\beta$ places non-negligible mass on near-optimal corrections. The argument assumes that the selector ranks candidates correctly; with a learned critic, a larger $M$ also offers more candidates whose cost the critic underestimates, so $M$ trades coverage against selection errors (Section~\ref{sec:sweep}). Unlike Drifting MPC, the selected candidate is executed through a model-based MPC, so dynamics and constraints are enforced by the optimiser rather than trusted from generated states. Algorithm~\ref{alg:train} summarises training.

\begin{algorithm}[t]
\caption{\method{} training}
\label{alg:train}
\begin{algorithmic}[1]
\Require contexts $\{c_n\}$, candidate set $\{y_j\}_{j=1}^K\subset\mathcal{Y}$, short MPC~\eqref{eq:mpc}
\For{each $c_n$ and $y_j$ \textbf{in parallel}}
  \State $J_{nj}\gets$ closed-loop cost of MPC with correction $y_j$ from $c_n$
\EndFor
\State $w_{nj}\propto \exp(-\beta\,\mathrm{clip}(\mathrm{zscore}_j(J_{nj})))$
\Repeat
  \State sample batch of contexts; $y_m\gets G_\phi(\epsilon_m,c_n)$, $m=1..M$
  \State compute $V(y_m)$ by~\eqref{eq:drift}; take a gradient step on~\eqref{eq:loss}
\Until{converged}
\State fit critic $\hat J_\psi$ on $\{(c_n,y_j,J_{nj})\}$ by least squares
\end{algorithmic}
\end{algorithm}

%% file: sec_exp_setup.tex
\section{Experiments}
\label{sec:exp}

\subsection{Setup}
\textbf{Vehicle and MPC.} We reuse the kinematic racing setting of ZipMPC~\cite{rickenbach2025zipmpc}: a Frenet-frame kinematic bicycle with $T=0.03$\,s, $l_f=l_r=0.05$\,m, $|a|\le 1$\,m/s$^2$, $|\delta|\le 0.4$\,rad, $v\le 1.8$\,m/s, lateral bound $|d|\le 0.17$\,m, and the manual contouring cost of~\cite{rickenbach2025zipmpc} (quadratic weights $3,1,0.01,0.01,0.01,1$ on $d,\varphi,v,\sigma_\Delta,a,\delta$ and linear progress weight $-8$). We add a friction limit $|v^2\tan\delta|/(l_f+l_r)\le 2$\,m/s$^2$ and a slower opponent on a straight track. The opponent drives at constant speed $v_o$ at lateral offset $d_o$; the MPC knows its predicted position and imposes an inflated elliptical keep-out constraint, while collisions in simulation use the true footprint. The short MPC uses $N_S=5$ (0.15\,s) and the long-horizon expert $N_L=30$ (0.9\,s); both are solved with IPOPT~\cite{wachter2006ipopt} through CasADi~\cite{andersson2019casadi} with hard constraints. If the short MPC is infeasible, the vehicle brakes fully.

\textbf{Contexts and task cost.} A context $c=(d_e,v_e,g,d_o,v_o)$ is sampled uniformly from $d_e\in[-0.05,0.05]$\,m, $v_e\in[1.2,1.8]$\,m/s, gap $g\in[0.6,1.2]$\,m, $d_o\in[-0.08,0.08]$\,m and $v_o\in[0.2,0.6]$\,m/s, so that the opponent is outside the short horizon but inside the long one. The correction is chosen at decision time and held for an 80-step (2.4\,s) closed-loop episode. $J$ is the negative progress normalised by $v_{\max}T\cdot 80$, plus $1$ if the vehicle collides or leaves the track. We use 800 contexts, the first 600 for training and the last 200 for testing. To check that results do not depend on this particular test set, the eight-dimensional experiments are also evaluated on two further sets of 1000 contexts drawn from the same distribution with different random seeds, never used for training or model selection.

\textbf{Statistics.} On the fresh contexts, the crash outcomes of two methods are paired by context and training seed and compared with an exact McNemar test; 95\% confidence intervals resample contexts. For each fresh set, the primary comparison was fixed in writing before the evaluation was run.

\textbf{Methods.} \emph{MPC$_{N_S}$} uses the manual cost ($\dC=0$); \emph{expert} is MPC$_{N_L}$ run in closed loop with two initial guesses (left and right) at the first step. \emph{ZipMPC-style} is a deterministic MLP regressing, for each context, the correction that minimises ZipMPC's imitation loss (weighted error of the first $N_D=5$ predicted states and inputs, with the weights of the released code) with respect to the expert's chosen solution; it is an idealised ZipMPC whose labels are exact per-context optima of its own loss. \emph{argmin reg.}\ regresses $\arg\min_y J(y;c)$; \emph{MSE tilted} regresses $\E_{p_\beta}[\dC\mid c]$. \emph{MDN} is a two-component mixture density network~\cite{bishop1994mdn} trained on the tilted weights. \emph{Critic search} minimises the learned critic $\hat J_\psi$ over a 201-point grid (one dimension) or over uniformly sampled corrections (eight dimensions), without a generator. \method{} uses $\beta=3$, $M=8$ candidates unless stated otherwise, $32$ negatives, temperatures $\{0.05,0.1,0.2\}$ in one dimension and $\{0.2,0.4,0.8\}$ in eight; all networks are three-layer MLPs with 256 units. ``M=1'' uses a single generated sample without selection; ``best-of-M'' selects with $\hat J_\psi$.

%% file: sec_exp_results.tex
\subsection{Is the optimal correction multimodal?}

\begin{figure}[t]
\centering
\includegraphics[width=\columnwidth]{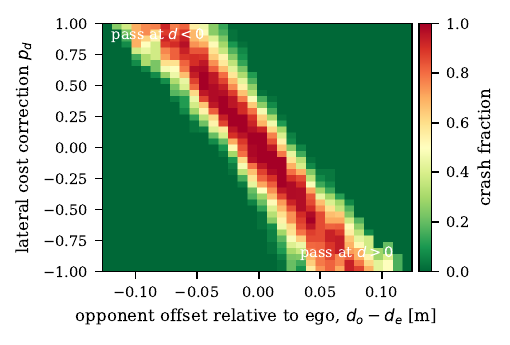}
\caption{Fraction of contexts in which a constant lateral correction $p_d$ leads to a crash, binned by the opponent's lateral offset relative to the ego vehicle. Good corrections lie on both sides of a moving crash band.}
\label{fig:landscape}
\end{figure}

Fig.~\ref{fig:landscape} shows, for all 800 contexts, which constant lateral corrections $p_d\in[-1,1]$ (41 values) lead to a crash. Crashes form a band whose position moves with the opponent's offset relative to the ego vehicle: corrections on one side of the band pre-position the car to pass at $d<0$, corrections on the other side to pass at $d>0$. The manual cost ($p_d=0$) crashes in 28.0\% of contexts, and 93.5\% of contexts contain at least one crashing correction. In 53.4\% of contexts both sides contain a non-crashing interval, i.e., the set of good corrections has two separated components. The long-horizon expert also has two distinct local solutions in 91\% of contexts, and in about half of those (51\%) their objective values differ by less than 0.5\% (median relative gap 0.48\%). Which mode the expert returns is therefore decided by small cost differences or, for a single-start solver, by the initial guess: a single IPOPT start picks the same side as the better of two starts in only 89.4\% of contexts.

\subsection{One-dimensional correction}

\begin{figure}[t]
\centering
\includegraphics[width=\columnwidth]{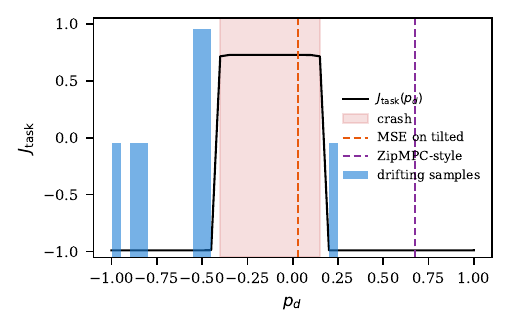}
\caption{A symmetric test context. The task cost $J$ is low on both sides of a crash interval. Drifting samples (histogram, $M=8$) cover both modes; the MSE solution on tilted labels lies inside the crash interval.}
\label{fig:example}
\end{figure}

Table~\ref{tab:main} reports closed-loop results with $p_d$ as the only learned parameter. Regressing the mean of the tilted distribution, which is what any MSE-trained map converges to when labels from both modes are mixed (Proposition~\ref{prop:avg}), crashes in 42.9\% of bimodal contexts: the predicted correction falls into the crash band (Fig.~\ref{fig:example}). \method{} with best-of-$M$ selection crashes in 0.2\% of all and 0.3\% of bimodal contexts, close to the oracle that picks the best of 41 evaluated corrections. Drawing a single sample without selection is worse (3.8\%), because samples near a mode's boundary are occasionally drawn.

Two points temper this result. First, the idealised ZipMPC-style regressor, trained on exact per-context optima of ZipMPC's imitation loss with respect to an expert that always returns the better of its two modes, is also nearly collision-free (0.8\%): its labels switch sides discontinuously but consistently with the context, and a sufficiently expressive network fits that switch. Second, in one dimension a learned critic with exhaustive grid search, or an MDN with the same best-of-$M$ selection, performs as well as \method.

\input{tab_1d}

\subsection{Sensitivity to expert labels}

Mode averaging returns as soon as the expert's mode choice is not a consistent function of the context. Table~\ref{tab:labels} retrains the ZipMPC-style regressor with labels from a single-start expert, as in a standard ZipMPC pipeline, and from an expert that breaks near-ties (relative gap below 0.5\%) at random, which models several near-equivalent demonstrators or solver restarts. Its crash rate rises from 0.5\% to 4.5\% and 16.5\% (31.6\% on bimodal contexts). \method{} does not use expert labels and stays at 0.5\%. The same holds for the other methods trained on closed-loop costs, which shows that much of the robustness comes from learning from cost evaluations rather than from imitation.

\input{tab_labels}

\subsection{Eight-dimensional stagewise correction}
\label{sec:hd}

To test whether the generator matters when exhaustive search is no longer available, we let the correction vary over the horizon, as in ZipMPC: $y\in[-1,1]^8$ sets the lateral weight $p_{d,i}=y_i$ and a progress-weight correction $\Delta p_{\sigma,i}=4y_{4+i}$ at stages $i=1,\dots,4$ (stage-0 terms are constant). For each of the same 800 contexts we evaluate the zero correction and $K=64$ uniformly sampled corrections in closed loop; on average 27.9\% of them crash. ZipMPC-style labels are the per-context minimisers of the imitation loss, found by bounded Powell search started from the best candidate, for the expert's chosen mode. Critic-based search now uses $M=8$ or 4096 uniformly sampled corrections.

Table~\ref{tab:hdext} reports closed-loop results on 1000 fresh test contexts for five training seeds, i.e., 5000 episodes per method. Ranking generated candidates with the critic gives the lowest crash rates among the learned methods: 0.74\% for \method{} and 0.76\% for the MDN. Their comparison was the primary test of this evaluation and shows no difference (exact McNemar $p=1$; difference $-0.02$ percentage points, 95\% CI $[-0.34,0.32]$). Both are clearly better than the idealised ZipMPC-style regressor (3.4\%; 6.2\% on bimodal contexts) and than ranking eight uniform samples with the same critic (2.1\%), with $p<10^{-8}$ in each of these comparisons, so the generated proposals, not the critic alone, carry the improvement. Ranking 4096 uniform samples is worse (3.8\%) than ranking eight: a large search finds corrections where the critic is wrong. With single-start or randomly tie-broken expert labels, the ZipMPC-style regressor degrades further (Table~\ref{tab:hdlabels}, measured on the original test contexts).

On the original 200 test contexts, \method{} with best-of-$M$ did not crash in any of 1000 episodes (five seeds), whereas the MDN crashed in 0.7\%. The fresh contexts show that this gap does not generalise: \method's crashes there occur in 29 distinct contexts, and a random set of 200 out of the 1000 contexts avoids all of them with probability 0.14\%. The original test set was simply favourable to \method.

\input{tab_hd_ext}
\input{tab_hd_labels}

\subsection{How many candidates are needed?}
\label{sec:sweep}

\begin{figure}[t]
\centering
\includegraphics[width=\columnwidth]{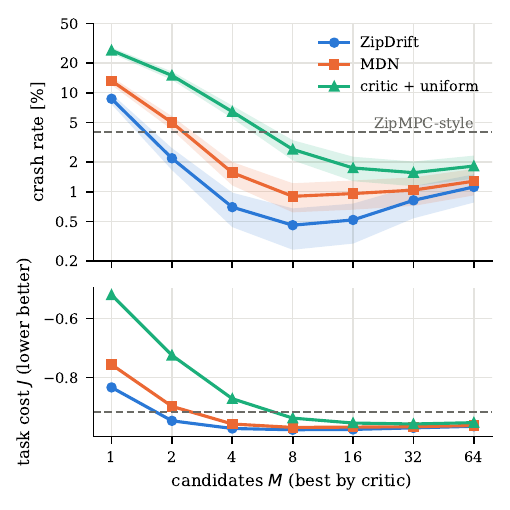}
\caption{Crash rate (log scale; shaded: 95\% bootstrap CI over contexts) and task cost against the number $M$ of candidates ranked by the critic, on 1000 unseen contexts and five training seeds. Dashed: ZipMPC-style regressor.}
\label{fig:sweep}
\end{figure}

Table~\ref{tab:hdext} also shows that single samples without selection crash much less often for \method{} (7.8\%) than for the MDN (14.9\%). Since this comparison was not planned, we tested it on a second set of 1000 unseen contexts, with the test fixed beforehand. For each context and seed, we draw 64 samples from each generator and let the critic select the best of the first $M$, for $M\in\{1,2,4,\dots,64\}$ (Fig.~\ref{fig:sweep}, Table~\ref{tab:sweep}). A single drifting sample crashes in 8.7\% of episodes against 13.2\% for the MDN ($p=1.4\cdot10^{-15}$). The gap persists for small budgets: with two candidates \method{} crashes in 2.2\% against 5.0\%, already below the ZipMPC-style regressor (4.0\% on these contexts), and it falls below 1\% with four candidates, where the MDN needs eight. Uniform samples ranked by the critic do not reach 1\% for any $M$. From eight candidates on, the two generators are close. The remaining differences at $M=8$ and $16$ are significant on this set but not on the first fresh set (0.54\% against 0.78\% at $M=8$, unadjusted $p=0.13$), and redrawing the candidate pools alone changes \method's crash rate at $M=8$ by about 0.2 percentage points, so we do not claim an advantage there. Selection remains essential: without it, a drifting sample crashes more often than the deterministic ZipMPC-style regressor.

\textbf{More candidates can hurt.} For both generators the crash rate is lowest at four to eight candidates and rises for larger $M$ (\method: 0.46\% at $M=8$, 1.12\% at $M=64$), and the task cost worsens slightly; ranked uniform samples also deteriorate after 16 to 32 candidates, and the first fresh set shows the same pattern. With more candidates, the critic's argmin increasingly selects corrections whose cost it underestimates. The best-of-$M$ argument of Section~\ref{sec:method} assumes an exact selector and does not capture this effect; in practice $M$ must be tuned, and better samples allow a smaller $M$.

\input{tab_sweep}

%% file: tab_1d.tex
\begin{table}[t]
\centering
\caption{One-dimensional correction $p_d$: closed-loop results on 200 test contexts (98 bimodal), mean$\pm$std over 3 training seeds; best-of-$M$ uses $M{=}8$.}
\label{tab:main}
\setlength{\tabcolsep}{3pt}
\footnotesize
\resizebox{\columnwidth}{!}{%
\begin{tabular}{lcccc}
\toprule
& \multicolumn{3}{c}{crash rate [\%] $\downarrow$} & \\ \cmidrule(lr){2-4}
Method & all & bimodal & unimodal & $J$ $\downarrow$ \\
\midrule
MPC$_{N_S}$ (manual cost) & 28.5$\pm$0.0 & 56.1$\pm$0.0 & 2.0$\pm$0.0 & -0.492$\pm$0.000 \\
Expert MPC$_{N_L}$ & 0.0$\pm$0.0 & 0.0$\pm$0.0 & 0.0$\pm$0.0 & -0.984$\pm$0.000 \\
Oracle (best of grid) & 0.0$\pm$0.0 & 0.0$\pm$0.0 & 0.0$\pm$0.0 & -0.984$\pm$0.000 \\
ZipMPC-style (deterministic) & 0.8$\pm$0.8 & 1.7$\pm$1.7 & 0.0$\pm$0.0 & -0.970$\pm$0.015 \\
argmin reg.\ (deterministic) & 1.5$\pm$0.8 & 3.1$\pm$1.7 & 0.0$\pm$0.0 & -0.958$\pm$0.014 \\
MSE tilted (deterministic) & 21.0$\pm$0.0 & 42.9$\pm$0.0 & 0.0$\pm$0.0 & -0.623$\pm$0.000 \\
Critic + grid search & 0.2$\pm$0.2 & 0.0$\pm$0.0 & 0.3$\pm$0.5 & -0.981$\pm$0.004 \\
MDN, $M{=}1$ & 4.0$\pm$0.4 & 3.4$\pm$0.5 & 4.6$\pm$0.5 & -0.915$\pm$0.007 \\
MDN, best-of-$M$ & 0.2$\pm$0.2 & 0.0$\pm$0.0 & 0.3$\pm$0.5 & -0.981$\pm$0.004 \\
\method, $M{=}1$ & 3.8$\pm$1.4 & 4.4$\pm$1.3 & 3.3$\pm$2.6 & -0.918$\pm$0.025 \\
\method, best-of-$M$ & 0.2$\pm$0.2 & 0.3$\pm$0.5 & 0.0$\pm$0.0 & -0.981$\pm$0.004 \\
\bottomrule
\end{tabular}}
\end{table}

%% file: tab_labels.tex
\begin{table}[t]
\centering
\caption{Effect of expert label consistency (seed 0). Crash rate [\%], all / bimodal contexts. Only the ZipMPC-style method uses expert labels.}
\label{tab:labels}
\setlength{\tabcolsep}{3pt}
\footnotesize
\begin{tabular}{lccc}
\toprule
Method & consistent & single-start & random tie \\
\midrule
ZipMPC-style (deterministic) & 0.5 / 1.0 & 4.5 / 8.2 & 16.5 / 31.6 \\
argmin reg.\ (deterministic) & 1.5 / 3.1 & 1.5 / 3.1 & 1.5 / 3.1 \\
MDN, best-of-$M$ & 0.0 / 0.0 & 0.0 / 0.0 & 0.5 / 0.0 \\
Critic + grid search & 0.5 / 0.0 & 0.5 / 0.0 & 0.5 / 0.0 \\
\method, best-of-$M$ & 0.5 / 1.0 & 0.5 / 1.0 & 0.5 / 1.0 \\
\bottomrule
\end{tabular}
\end{table}

%% file: tab_hd_ext.tex
\begin{table}[t]
\centering
\caption{Eight-dimensional stagewise correction: closed-loop results on 1000 fresh test contexts (549 bimodal), mean$\pm$std over 5 training seeds; best-of-$M$ uses $M{=}8$.}
\label{tab:hdext}
\setlength{\tabcolsep}{3pt}
\footnotesize
\resizebox{\columnwidth}{!}{%
\begin{tabular}{lcccc}
\toprule
& \multicolumn{3}{c}{crash rate [\%] $\downarrow$} & \\ \cmidrule(lr){2-4}
Method & all & bimodal & unimodal & $J$ $\downarrow$ \\
\midrule
MPC$_{N_S}$ (manual cost) & 28.30$\pm$0.00 & 50.64$\pm$0.00 & 1.11$\pm$0.00 & -0.496$\pm$0.000 \\
Expert MPC$_{N_L}$ & 0.00$\pm$0.00 & 0.00$\pm$0.00 & 0.00$\pm$0.00 & -0.985$\pm$0.000 \\
ZipMPC-style (deterministic) & 3.42$\pm$0.53 & 6.23$\pm$0.97 & 0.00$\pm$0.00 & -0.927$\pm$0.009 \\
argmin reg.\ (deterministic) & 5.74$\pm$0.30 & 10.42$\pm$0.48 & 0.04$\pm$0.09 & -0.886$\pm$0.005 \\
MSE tilted (deterministic) & 12.16$\pm$0.21 & 22.11$\pm$0.38 & 0.04$\pm$0.09 & -0.777$\pm$0.004 \\
Critic + $M$ random & 2.12$\pm$0.62 & 3.35$\pm$1.04 & 0.62$\pm$0.41 & -0.947$\pm$0.011 \\
Critic + 4096 random & 3.76$\pm$1.12 & 3.61$\pm$0.92 & 3.95$\pm$1.53 & -0.918$\pm$0.020 \\
MDN, $M{=}1$ & 14.86$\pm$0.96 & 24.63$\pm$2.02 & 2.97$\pm$0.54 & -0.729$\pm$0.016 \\
MDN, best-of-$M$ & 0.76$\pm$0.26 & 1.06$\pm$0.51 & 0.40$\pm$0.09 & -0.971$\pm$0.004 \\
\method, $M{=}1$ & 7.78$\pm$0.92 & 13.26$\pm$1.61 & 1.11$\pm$0.14 & -0.851$\pm$0.016 \\
\method, best-of-$M$ & 0.74$\pm$0.41 & 0.91$\pm$0.69 & 0.53$\pm$0.11 & -0.972$\pm$0.007 \\
\bottomrule
\end{tabular}}
\end{table}

%% file: tab_hd_labels.tex
\begin{table}[t]
\centering
\caption{Eight-dimensional correction: effect of expert label consistency (seed 0, original 200 test contexts). Crash rate [\%], all / bimodal.}
\label{tab:hdlabels}
\setlength{\tabcolsep}{3pt}
\footnotesize
\begin{tabular}{lccc}
\toprule
Method & consistent & single-start & random tie \\
\midrule
ZipMPC-style (deterministic) & 3.0 / 6.1 & 7.5 / 14.3 & 16.0 / 26.5 \\
argmin reg.\ (deterministic) & 7.5 / 15.3 & 7.5 / 15.3 & 7.5 / 15.3 \\
Critic + $M$ random & 1.0 / 2.0 & 1.0 / 2.0 & 0.5 / 1.0 \\
MDN, best-of-$M$ & 0.5 / 1.0 & 0.5 / 1.0 & 0.0 / 0.0 \\
\method, best-of-$M$ & 0.0 / 0.0 & 0.0 / 0.0 & 0.0 / 0.0 \\
\bottomrule
\end{tabular}
\end{table}

%% file: tab_sweep.tex
\begin{table}[t]
\centering
\caption{Crash rate [\%] against the number of critic-ranked candidates $M$ on 1000 further unseen contexts, 5000 episodes per entry (ZipMPC-style: 4.04). $\Delta$: \method{} minus MDN with 95\% bootstrap CI over contexts; $p$: exact McNemar, Holm-adjusted for $M>1$; $M{=}1$ is the primary test.}
\label{tab:sweep}
\setlength{\tabcolsep}{3pt}
\footnotesize
\resizebox{\columnwidth}{!}{%
\begin{tabular}{rccccc}
\toprule
$M$ & \method & MDN & Critic + uniform & $\Delta$ & $p$ \\
\midrule
1 & 8.70 & 13.22 & 26.80 & -4.52 [-5.70, -3.32] & $<10^{-14}$ \\
2 & 2.18 & 5.00 & 14.94 & -2.82 [-3.52, -2.14] & $<10^{-15}$ \\
4 & 0.70 & 1.56 & 6.44 & -0.86 [-1.28, -0.46] & $<10^{-4}$ \\
8 & 0.46 & 0.90 & 2.68 & -0.44 [-0.76, -0.14] & 0.02 \\
16 & 0.52 & 0.96 & 1.74 & -0.44 [-0.76, -0.12] & 0.02 \\
32 & 0.82 & 1.04 & 1.56 & -0.22 [-0.54, +0.10] & 0.47 \\
64 & 1.12 & 1.28 & 1.82 & -0.16 [-0.52, +0.20] & 0.47 \\
\bottomrule
\end{tabular}}
\end{table}

%% file: sec_discussion.tex
\section{Discussion and Limitations}
\label{sec:discussion}

\textbf{What the evidence supports.} Multimodal optimal corrections are common in the overtaking task, and deterministic imitation is only as reliable as the consistency of the expert's mode choice. Learning from closed-loop cost evaluations removes that dependence, and ranking generated candidates with a critic is clearly better than deterministic maps and than ranking random candidates with the same budget. The evidence for drifting in particular is narrower. In one dimension a critic with grid search or an MDN performs as well as \method, and in eight dimensions the MDN reaches the same crash rate once eight candidates are ranked. What drifting improves is the quality of individual samples: it needs four candidates instead of eight to fall below a 1\% crash rate. Here candidates are scored by a cheap critic, so this matters mainly where scoring is expensive, for example when each candidate is evaluated by solving the short-horizon MPC, or where no reliable critic is available. We have tested neither case, nor compared other few-step generators such as flow-matching or consistency models.

\textbf{Test sets.} Results on the original 200 test contexts were more favourable to \method{} than results on fresh contexts (Section~\ref{sec:hd}). The one-dimensional experiments and both label studies (Tables~\ref{tab:main}, \ref{tab:labels} and~\ref{tab:hdlabels}) use only those contexts and should be read with this in mind.

\textbf{Scope.} The benchmark is deliberately minimal: a kinematic bicycle on a straight track, one opponent at constant speed, a correction chosen at decision time and held for the episode, and simulation only. Random corrections pass in most contexts, so the eight-dimensional search problem is not hard. We have not evaluated full laps on curved tracks, the Pacejka model, more opponents, or hardware.

\textbf{Receding-horizon use.} ZipMPC recomputes the correction at every step. Doing so with a stochastic generator risks switching between modes from one step to the next; reusing the noise $\epsilon$ across steps, or conditioning on the previously selected mode, are natural remedies that remain to be tested.

\textbf{Guarantees.} The Gibbs characterisation and the best-of-$M$ argument concern the choice of a correction, not closed-loop stability or recursive feasibility. Terminal ingredients in the short MPC are compatible with \method{} and are needed for formal guarantees.

\textbf{Labelling cost.} Each training context required 65 closed-loop simulations, more solver calls per context than differentiable-MPC training, although they are trivially parallel and use the deployment solver; labelling 800 contexts took 18 minutes on 28 CPU cores. Sampling candidates adaptively around the current generator would reduce this cost.

%% file: sec_conclusion.tex
\section{Conclusion}
\label{sec:conclusion}

Compressing a long MPC horizon into a learned short-horizon cost requires representing decisions, such as the side on which to overtake, that the short horizon cannot see. Such decisions make the optimal cost correction multimodal and make imitation of a long-horizon expert sensitive to how the expert breaks near-ties. \method{} replaces the deterministic context-to-cost map by a cost-tilted one-step drifting generator in the space of cost corrections, trained only from closed-loop cost evaluations and combined with candidate selection and a model-based MPC. In a minimal overtaking benchmark, critic-ranked generated corrections reduced the crash rate from 3.4\% for an idealised deterministic ZipMPC-style map to below 1\%. A mixture density network does equally well with eight candidates; drifting produces better individual samples and needs fewer candidates, while too many candidates let the selection exploit errors of the critic. Next steps are receding-horizon use with mode persistence, curved tracks with the Pacejka model, higher-dimensional time-varying corrections, comparisons with other few-step generators, selection that is robust to critic errors, and hardware experiments.

\section*{Acknowledgment}
Parts of the code and text of this work were drafted with the assistance of an AI system (Claude, Anthropic). The author reviewed all content and takes full responsibility for it.

%% file: main.bbl
\begin{thebibliography}{10}
\providecommand{\url}[1]{#1}
\csname url@samestyle\endcsname
\providecommand{\newblock}{\relax}
\providecommand{\bibinfo}[2]{#2}
\providecommand{\BIBentrySTDinterwordspacing}{\spaceskip=0pt\relax}
\providecommand{\BIBentryALTinterwordstretchfactor}{4}
\providecommand{\BIBentryALTinterwordspacing}{\spaceskip=\fontdimen2\font plus
\BIBentryALTinterwordstretchfactor\fontdimen3\font minus
  \fontdimen4\font\relax}
\providecommand{\BIBforeignlanguage}[2]{{%
\expandafter\ifx\csname l@#1\endcsname\relax
\typeout{** WARNING: IEEEtran.bst: No hyphenation pattern has been}%
\typeout{** loaded for the language `#1'. Using the pattern for}%
\typeout{** the default language instead.}%
\else
\language=\csname l@#1\endcsname
\fi
#2}}
\providecommand{\BIBdecl}{\relax}
\BIBdecl

\bibitem{rickenbach2025zipmpc}
R.~Rickenbach, A.~Lahoud, E.~Schaffernicht, M.~Zeilinger, and J.~A. Stork,
  ``{ZipMPC}: Compressed context-dependent {MPC} cost via imitation learning,''
  in \emph{Proceedings of the 9th Conference on Robot Learning (CoRL)}, ser.
  Proceedings of Machine Learning Research, vol. 305.\hskip 1em plus 0.5em
  minus 0.4em\relax PMLR, 2025, pp. 3115--3136.

\bibitem{trisovic2026vigor}
J.~Trisovic, R.~Rickenbach, M.~Landert, P.~Minos, and M.~N. Zeilinger,
  ``{VIGOR-MPC}: Vision-guided objective regulation {MPC},'' in
  \emph{Conference on Robot Learning (CoRL)}, 2026, to appear.
  arXiv:2610.20333.

\bibitem{frohlich2022contextual}
L.~P. Fr{\"o}hlich, C.~K{\"u}ttel, E.~Arcari, L.~Hewing, M.~N. Zeilinger, and
  A.~Carron, ``Contextual tuning of model predictive control for autonomous
  racing,'' in \emph{2022 IEEE/RSJ International Conference on Intelligent
  Robots and Systems (IROS)}, 2022, pp. 10\,555--10\,562.

\bibitem{zarrouki2026sgrl}
B.~Zarrouki, A.~Thobani, J.~Hoffmann, M.~Piccinini, R.~Reiter, F.~Jahncke,
  S.~Gros, D.~Scaramuzza, and J.~Betz, ``Accelerating reinforcement learning
  via {MPC} solver-gradient guidance for weights-varying {MPC},'' 2026,
  arXiv:2609.01061.

\bibitem{deng2026drifting}
M.~Deng, H.~Li, T.~Li, Y.~Du, and K.~He, ``Generative modeling via drifting,''
  in \emph{Advances in Neural Information Processing Systems (NeurIPS)}, 2026,
  to appear. arXiv:2602.04770.

\bibitem{foffano2026drifting}
D.~Foffano, A.~Russo, and A.~Proutiere, ``Receding-horizon control via drifting
  models,'' 2026, arXiv:2604.04528.

\bibitem{puthumanaillam2026kdp}
G.~Puthumanaillam and M.~Ornik, ``Amortizing trajectory diffusion with keyed
  drift fields,'' 2026, arXiv:2603.14056.

\bibitem{gruene2009horizon}
L.~Gr{\"u}ne, ``Analysis and design of unconstrained nonlinear {MPC} schemes
  for finite and infinite dimensional systems,'' \emph{SIAM Journal on Control
  and Optimization}, vol.~48, no.~2, pp. 1206--1228, 2009.

\bibitem{jain2021costdesign}
A.~Jain, L.~Chan, D.~S. Brown, and A.~D. Dragan, ``Optimal cost design for
  model predictive control,'' in \emph{Proceedings of the 3rd Conference on
  Learning for Dynamics and Control (L4DC)}, ser. Proceedings of Machine
  Learning Research, vol. 144.\hskip 1em plus 0.5em minus 0.4em\relax PMLR,
  2021, pp. 1205--1217.

\bibitem{liniger2015mpcc}
A.~Liniger, A.~Domahidi, and M.~Morari, ``Optimization-based autonomous racing
  of 1:43 scale {RC} cars,'' \emph{Optimal Control Applications and Methods},
  vol.~36, no.~5, pp. 628--647, 2015.

\bibitem{amos2018diffmpc}
B.~Amos, I.~Jimenez, J.~Sacks, B.~Boots, and J.~Z. Kolter, ``Differentiable
  {MPC} for end-to-end planning and control,'' in \emph{Advances in Neural
  Information Processing Systems (NeurIPS)}, vol.~31.\hskip 1em plus 0.5em
  minus 0.4em\relax Curran Associates, Inc., 2018.

\bibitem{gros2020dempc}
S.~Gros and M.~Zanon, ``Data-driven economic {NMPC} using reinforcement
  learning,'' \emph{IEEE Transactions on Automatic Control}, vol.~65, no.~2,
  pp. 636--648, 2020.

\bibitem{reiter2024ac4mpc}
R.~Reiter, A.~Ghezzi, K.~Baumg{\"a}rtner, J.~Hoffmann, R.~D. McAllister, and
  M.~Diehl, ``{AC4MPC}: Actor-critic reinforcement learning for guiding model
  predictive control,'' \emph{IEEE Transactions on Control Systems Technology},
  vol.~34, no.~1, pp. 395--410, 2026.

\bibitem{hertneck2018approxmpc}
M.~Hertneck, J.~K{\"o}hler, S.~Trimpe, and F.~Allg{\"o}wer, ``Learning an
  approximate model predictive controller with guarantees,'' \emph{IEEE Control
  Systems Letters}, vol.~2, no.~3, pp. 543--548, 2018.

\bibitem{janner2022diffuser}
M.~Janner, Y.~Du, J.~Tenenbaum, and S.~Levine, ``Planning with diffusion for
  flexible behavior synthesis,'' in \emph{Proceedings of the 39th International
  Conference on Machine Learning (ICML)}, ser. Proceedings of Machine Learning
  Research, vol. 162.\hskip 1em plus 0.5em minus 0.4em\relax PMLR, 2022, pp.
  9902--9915.

\bibitem{chi2023diffusionpolicy}
C.~Chi, S.~Feng, Y.~Du, Z.~Xu, E.~Cousineau, B.~Burchfiel, and S.~Song,
  ``Diffusion policy: Visuomotor policy learning via action diffusion,'' in
  \emph{Robotics: Science and Systems XIX}, 2023.

\bibitem{zhou2024dmpc}
\BIBentryALTinterwordspacing
G.~Zhou, S.~Swaminathan, R.~V. Raju, J.~S. Guntupalli, W.~Lehrach, J.~Ortiz,
  A.~Dedieu, M.~L{\'a}zaro-Gredilla, and K.~Murphy, ``Diffusion model
  predictive control,'' \emph{Transactions on Machine Learning Research}, 2025.
  [Online]. Available: \url{https://openreview.net/forum?id=pvtgffHtJm}
\BIBentrySTDinterwordspacing

\bibitem{pan2024mbd}
C.~Pan, Z.~Yi, G.~Shi, and G.~Qu, ``Model-based diffusion for trajectory
  optimization,'' in \emph{Advances in Neural Information Processing Systems
  (NeurIPS)}, vol.~37.\hskip 1em plus 0.5em minus 0.4em\relax Curran
  Associates, Inc., 2024, pp. 57\,914--57\,943.

\bibitem{julbe2025diffampc}
P.~M. Julbe, J.~Nubert, H.~Hose, S.~Trimpe, and K.~J. Kuchenbecker,
  ``Diffusion-based approximate {MPC}: Fast and consistent imitation of
  multi-modal action distributions,'' in \emph{2025 IEEE/RSJ International
  Conference on Intelligent Robots and Systems (IROS)}, 2025, pp. 5633--5640.

\bibitem{ma2025codig}
H.~Ma, S.~Bodmer, A.~Carron, M.~Zeilinger, and M.~Muehlebach,
  ``Constraint-aware diffusion guidance for robotics: Real-time obstacle
  avoidance for autonomous racing,'' in \emph{Proceedings of the 9th Conference
  on Robot Learning (CoRL)}, ser. Proceedings of Machine Learning Research,
  vol. 305.\hskip 1em plus 0.5em minus 0.4em\relax PMLR, 2025, pp. 1756--1776.

\bibitem{lu2025diffplanner}
\BIBentryALTinterwordspacing
H.~Lu, D.~Han, Y.~Shen, and D.~Li, ``What makes a good diffusion planner for
  decision making?'' in \emph{The Thirteenth International Conference on
  Learning Representations (ICLR)}, 2025. [Online]. Available:
  \url{https://openreview.net/forum?id=7BQkXXM8Fy}
\BIBentrySTDinterwordspacing

\bibitem{zhang2025efm}
\BIBentryALTinterwordspacing
S.~Zhang, W.~Zhang, and Q.~Gu, ``Energy-weighted flow matching for offline
  reinforcement learning,'' in \emph{The Thirteenth International Conference on
  Learning Representations (ICLR)}, 2025. [Online]. Available:
  \url{https://openreview.net/forum?id=HA0oLUvuGI}
\BIBentrySTDinterwordspacing

\bibitem{levine2018controlinference}
S.~Levine, ``Reinforcement learning and control as probabilistic inference:
  Tutorial and review,'' 2018, arXiv:1805.00909.

\bibitem{williams2017mppi}
G.~Williams, N.~Wagener, B.~Goldfain, P.~Drews, J.~M. Rehg, B.~Boots, and E.~A.
  Theodorou, ``Information theoretic {MPC} for model-based reinforcement
  learning,'' in \emph{2017 IEEE International Conference on Robotics and
  Automation (ICRA)}, 2017, pp. 1714--1721.

\bibitem{xia2026dfp}
J.~Koo, M.~Park, J.~Choi, Y.~Min, and M.~Sung, ``Drifting field policy: A
  one-step generative policy via {W}asserstein gradient flow,'' 2026,
  arXiv:2605.07727.

\bibitem{wachter2006ipopt}
A.~W{\"a}chter and L.~T. Biegler, ``On the implementation of an interior-point
  filter line-search algorithm for large-scale nonlinear programming,''
  \emph{Mathematical Programming}, vol. 106, no.~1, pp. 25--57, 2006.

\bibitem{andersson2019casadi}
J.~A.~E. Andersson, J.~Gillis, G.~Horn, J.~B. Rawlings, and M.~Diehl,
  ``{CasADi}: a software framework for nonlinear optimization and optimal
  control,'' \emph{Mathematical Programming Computation}, vol.~11, no.~1, pp.
  1--36, 2019.

\bibitem{bishop1994mdn}
C.~M. Bishop, ``Mixture density networks,'' Neural Computing Research Group,
  Aston University, Birmingham, U.K., Technical Report NCRG/94/004, Feb. 1994.

\end{thebibliography}
